\documentclass[11pt,a4paper]{article}
\usepackage[hyperref]{emnlp2020}
\usepackage{times}
\usepackage{latexsym}
\usepackage{bm}
\usepackage{amsmath}
\usepackage{amssymb}
\usepackage{amsfonts}
\usepackage{booktabs}
\usepackage{graphicx}
\usepackage{csquotes}
\usepackage{colortbl}
\usepackage{tikz,pgfplots}
\usetikzlibrary{shapes.geometric}
\usetikzlibrary{arrows.meta,arrows}
\usepackage[ruled,vlined]{algorithm2e}
\renewcommand{\UrlFont}{\ttfamily\small}
\usepackage{enumitem}
\usepackage{arydshln}
\usepackage{microtype}

\aclfinalcopy 
\def\aclpaperid{2511} 

\DeclareMathOperator*{\argmin}{\arg\!\min}

\title{\textsc{PAIR}: Planning and Iterative Refinement in Pre-trained Transformers \\for Long Text Generation}

\author{Xinyu Hua \\
  Khoury College of Computer Sciences \\
  Northeastern University \\
  Boston, MA \\
  \texttt{hua.x@northeastern.edu} \\\And
  Lu Wang \\
  Computer Science and Engineering \\
  University of Michigan \\
  Ann Arbor, MI \\
  \texttt{wangluxy@umich.edu} \\}

\date{}

\begin{document}
\maketitle
\begin{abstract}
Pre-trained Transformers have enabled impressive breakthroughs in generating long and fluent text, yet their outputs are often ``rambling" without coherently arranged content. 
In this work, we present a novel content-controlled text generation framework, \textbf{\textsc{PAIR}}, with \underline{p}lanning \underline{a}nd \underline{i}terative \underline{r}efinement, which is built upon a large model, BART. 
We first adapt the BERT model to automatically construct the content plans, consisting of keyphrase assignments and their corresponding sentence-level positions. 
The BART model is employed for generation without modifying its structure.
We then propose a refinement algorithm to gradually enhance the generation quality within the sequence-to-sequence framework.
Evaluation with automatic metrics shows that adding planning consistently improves the generation quality on three distinct domains, with an average of 20 BLEU points and 12 METEOR points improvements. In addition, human judges rate our system outputs to be more relevant and coherent than comparisons without planning. 
\end{abstract}

\section{Introduction}
\label{sec:intro}
\begin{figure}[t]
    \hspace{-1mm}
    \includegraphics[width=77mm]{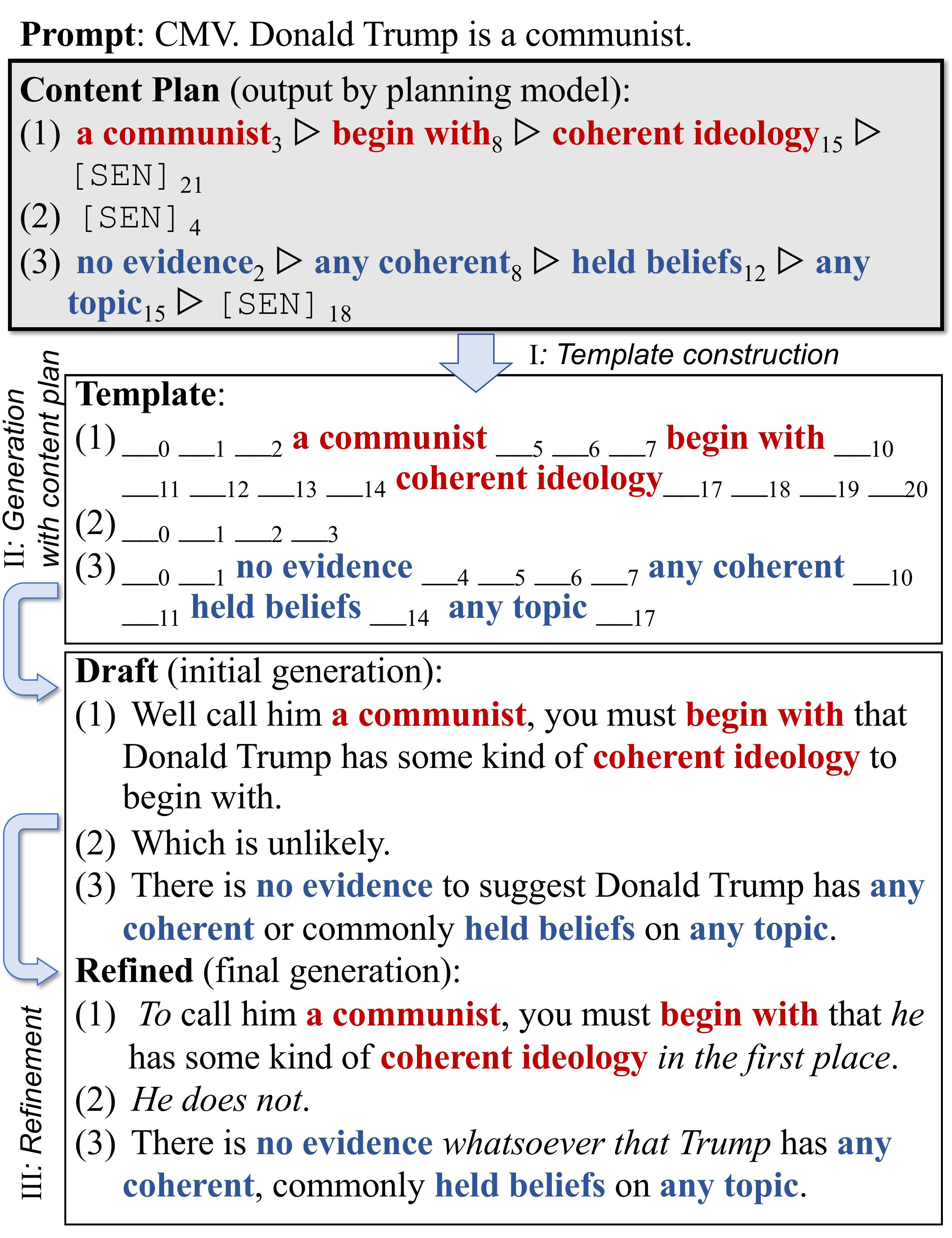}
    \caption{
An argument generation example using Reddit ChangeMyView. 
[Top] Partial output by our planner with keyphrase assignment and positions (in subscripts) for each sentence, segmented by special token \texttt{[SEN]}, from which a template is constructed.
[Bottom] A draft is first produced and then refined, with updated words highlighted in \textit{italics}. 
}
\label{fig:intro-example}
\end{figure}

Large pre-trained language models are the cornerstone of many state-of-the-art models in various natural language understanding and generation tasks~\cite{devlin-etal-2019-bert,liu2019roberta,lewis-etal-2020-bart}, yet they are far from perfect.
In generation tasks, although models like GPT-2~\cite{radford2019language} are able to produce plausible text, their spontaneous nature limits their utility in actual applications, e.g., users cannot specify what contents to include, and in what order.

To make large models more useful in practice, and to improve their generation quality, we believe it is critical to inform them of {\it when to say what}, which is addressed as {\it content planning} in traditional generation systems~\cite{duboue-mckeown-2001-empirically,stent-etal-2004-trainable}. 
Specially designed control codes and auxiliary planning modules have been integrated into neural models~\cite{keskar2019ctrl,moryossef-etal-2019-step,hua-wang-2019-sentence}, yet those solutions require model architecture modification or retraining, making text generation with large models a very costly endeavor. 

To this end, this work aims to bring new insights into {\it how to effectively incorporate content plans into large models to generate more relevant and coherent text}. 
We first study {\it a planning model trained from BERT}~\cite{devlin-etal-2019-bert} to produce the initial content plan, which assigns keyphrases to different sentences and predicts their positions. 
Next, we propose a {\it content-controlled text generation framework}, built upon the pre-trained sequence-to-sequence (seq2seq) Transformer model BART~\cite{lewis-etal-2020-bart}. 
As shown in Figure~\ref{fig:intro-example}, our generation model takes in a content plan consisting of \textit{keyphrase assignments} and their corresponding \textit{positions} for each sentence.
The plan is encoded as a template, with \texttt{[MASK]} tokens added at positions where no content is specified. 
Our model then outputs a fluent and coherent multi-sentence text (draft) to reflect the plan. 
This is done by fine-tuning BART without modifying its architecture. 

Furthermore, we present an {\it iterative refinement algorithm} to improve the generation in multiple passes, within the seq2seq framework. At each iteration, tokens with low generation confidence are replaced with \texttt{[MASK]} to compose a new template, from which a new output is produced. Unlike prior refinement algorithms that only permit editing in place, our solution offers more flexibility.  Figure~\ref{fig:intro-example} exemplifies the refinement outcome.

We call our system \textbf{\textsc{PAIR}} (Planning And Iterative Refinement).\footnote{Code and data are available at: \url{http://xinyuhua.github.io/Resources/emnlp20/}}
It is experimented on three distinct domains: counter-argument generation with Reddit ChangeMyView data, opinion article writing with the New York Times (NYT) corpus\footnote{\url{https://catalog.ldc.upenn.edu/LDC2008T19}}~\cite{sandhaus2008new}, and news report production on NYT. 
Automatic evaluation with BLEU, ROUGE, and METEOR shows that, by informing the generation model with sentence-level content plans, our model significantly outperforms a BART model fine-tuned with the same set of keyphrases as input (\S~\ref{sec:auto-eval}). Human judges also rate our system outputs as more relevant and coherent (\S~\ref{sec:human}). 
Additionally, our iterative refinement strategy consistently improves the generation quality according to both automatic scores and human evaluation. 
Finally, our model achieves better content control by reflecting the specified keyphrases in the content plan, whose outputs are preferred by human to another version with weaker control. 

To summarize, our major contributions include:

$\bullet$ We propose a novel content planner built upon BERT to facilitate long-form text generation.

$\bullet$ We present a novel template mask-and-fill method to incorporate content planning into generation models based on BART.

$\bullet$ We devise an iterative refinement algorithm that works within the seq2seq framework to flexibly improve the generation quality.

\section{Related Work}
\label{sec:related}
\noindent \textbf{Content Planning as a Generation Component.} 
Despite the impressive progress made in many generation tasks, neural systems are known to produce low-quality content~\cite{wiseman-etal-2017-challenges,rohrbach-etal-2018-object}, often with low relevance~\cite{li-etal-2016-diversity} and poor discourse structure~\cite{zhao-etal-2017-learning,xu-etal-2020-discourse}. 
Consequently, planning modules are designed and added into neural systems to enhance content relevance~\cite{wiseman-etal-2018-learning,moryossef-etal-2019-step,yao2019plan,hua-wang-2019-sentence}.
However, it is still an open question to include content plans in large models, given the additional and expensive model retraining required.
This work innovates by adding content plans as masked templates and designing refinement strategy to further boost generation performance, without architectural change.

\smallskip
\noindent \textbf{Controlled Text Generation.}
Our work is also in line with the study of controllability of neural text generation models.
This includes manipulating the syntax~\cite{dusek-jurcicek-2016-sequence, goyal-durrett-2020-neural}
and semantics~\cite{wen-etal-2015-semantically, chen-etal-2019-multi} of the output. Specific applications encourage the model to cover a given topic~\cite{wang-etal-2017-steering,see-etal-2019-makes}, mention specified entities~\cite{fan-etal-2018-controllable}, or display a certain attribute~\cite{hu2017toward,luo-etal-2019-learning,balakrishnan-etal-2019-constrained}. 
However, most existing work relies on model engineering, limiting the generalizability to new domains and adaptability to large pre-trained Transformers. 
One exception is the Plug and Play model~\cite{Dathathri2020Plug}, which directly modifies the key and value states of GPT-2~\cite{radford2019language}. 
However, since the signal is derived from the whole generated text, it is too coarse to provide precise sentence-level content control. 
Here, we instead gain fine-grained controllability through keyphrase assignment and positioning per sentence, which can be adapted to any off-the-shelf pre-trained Transformer generators.

\smallskip
\noindent \textbf{Iterative Refinement} has been studied in machine translation~\cite{lee-etal-2018-deterministic,freitag-etal-2019-ape,mansimov2019generalized,Kasai2020DisCo} to gradually improve translation quality. 
Refinement is also used with masked language models 
to improve fluency of non-autoregressive generation outputs~\cite{ghazvininejad-etal-2019-mask, lawrence-etal-2019-attending}. 
Our work uses BART~\cite{lewis-etal-2020-bart}, a state-of-the-art seq2seq model that 
offers better generalizability and stronger capacity for long text generation. 
Our proposed strategy substantially differs from prior solutions that rely on in-place word substitutions~\cite{novak2016iterative,xia2017deliberation,weston-etal-2018-retrieve}, as we leverage the seq2seq architecture to offer more flexible edits.

\section{Content-controlled Text Generation with \textsc{PAIR}}
\label{sec:model}
\noindent \textbf{Task Description.} 
Our input consists of (1) a sentence-level prompt $\bm{x}$, such as a news headline, or a proposition in an argument, and (2) a set of keyphrases $\bm{m}$ that are relevant to the prompt. 
The system aims to generate $\bm{y}$ that contains multiple sentences, as in a news report or an argument, by reflecting the keyphrases in a coherent way. 

In this section, we first introduce content planning built upon BERT, that assigns keyphrases into sentences and predicts their positions (\S~\ref{sec:planning}). Then we propose a seq2seq generation framework with BART fine-tuning that includes a given content plan derived from keyphrases $\bm{m}$ (\S~\ref{sec:realization}). 
Finally, \S~\ref{sec:refinement} discusses improving generation quality by iteratively masking the less confident predictions and regenerating within our framework. 

\subsection{Content Planning with BERT}
\label{sec:planning}

\begin{figure}[t]
    \centering
    \includegraphics[width=78mm]{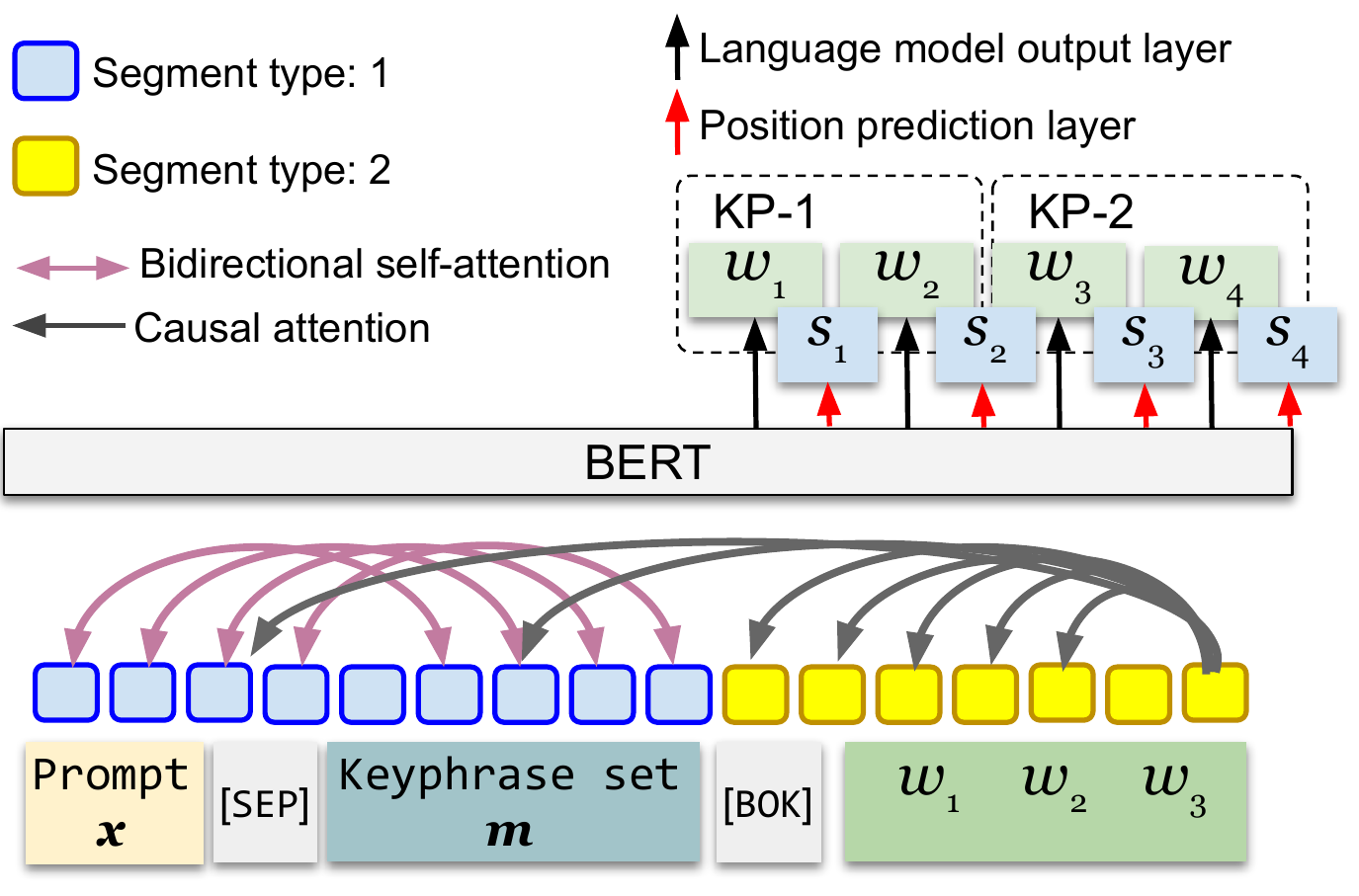}
    \caption{
    Content planning with BERT. We use bidirectional self-attentions for input encoding, and apply causal self-attentions for keyphrase assignment and position prediction. The input ($\bm{x}$, $\bm{m}$) and output keyphrase assignments ($\bm{m}'$) are distinguished by different segment embeddings. 
    }
    \label{fig:bert-planner}
\end{figure}

Our content planner is trained from BERT to assign keyphrases to different sentences and predict their corresponding positions. 
As shown in Figure~\ref{fig:bert-planner}, the concatenation of prompt $\bm{x}$ and unordered keyphrases $\bm{m}$ 
is encoded with bidirectional self-attentions. 
Keyphrase assignments are produced autoregressively as a sequence of tokens $\bm{m}'=\{w_j\}$, with their positions in the sentence $\bm{s}=\{s_j\}$ predicted as a sequence tagging task. 

We choose BERT because it has been shown to be effective at both language modeling and sequence tagging. Moreover, we leverage its segment embedding to distinguish the input and output sequences. 
Specifically, we reuse its pre-trained language model output layer 
for keyphrase assignment. We further design a separate keyphrase positioning layer to predict token position $s_j$ as the relative distance from each sentence's beginning: 

{
\setlength{\abovedisplayskip}{2pt}
\begin{align}
    p(s_j|\bm{w}_{\leq j}) = \text{softmax}(\bm{H}^L\bm{W}_{s}) \label{eq:offset_prediction}
\end{align}
}
%
where $\bm{H}^L$ is the last layer hidden states of the Transformer,
and $\bm{W}_s$ are the newly added keyphrase positioning parameters learned during BERT fine-tuning. 
The range of allowed positions is from $0$ to $127$. 

Noticeably, as our prediction is done autoregressively, attentions should only consider the generated tokens, but not the future tokens. 
However, BERT relies on bidirectional self-attentions to attend to both left and right. To resolve this discrepancy, we apply causal attention masks~\cite{dong2019unified} over $\bm{m}'$ to disallow attending to the future (gray arrows in Figure~\ref{fig:bert-planner}). 

\smallskip
\noindent {\bf Training the Planner.} We extract keyphrases and acquire their ground-truth positions from human-written references, and fine-tune BERT with cross-entropy losses for both assignment and positioning, with a scaling factor $0.1$ over the positioning loss. 

\smallskip
\noindent {\bf Inference.} 
A \texttt{[BOK]} token signals the beginning of keyphrase assignment generation. 
We employ a greedy decoding algorithm, and limit the output vocabulary to tokens in $\bm{m}$ and ensure each keyphrase is generated at most once. 
To allow sentence-level content planning, a special \texttt{[SEN]} token is generated to represent the sentence boundary, with its predicted position indicating the length. 
The planning process terminates when \texttt{[EOS]} is produced.

\begin{figure*}[t]
    \centering
    \includegraphics[width=160mm]{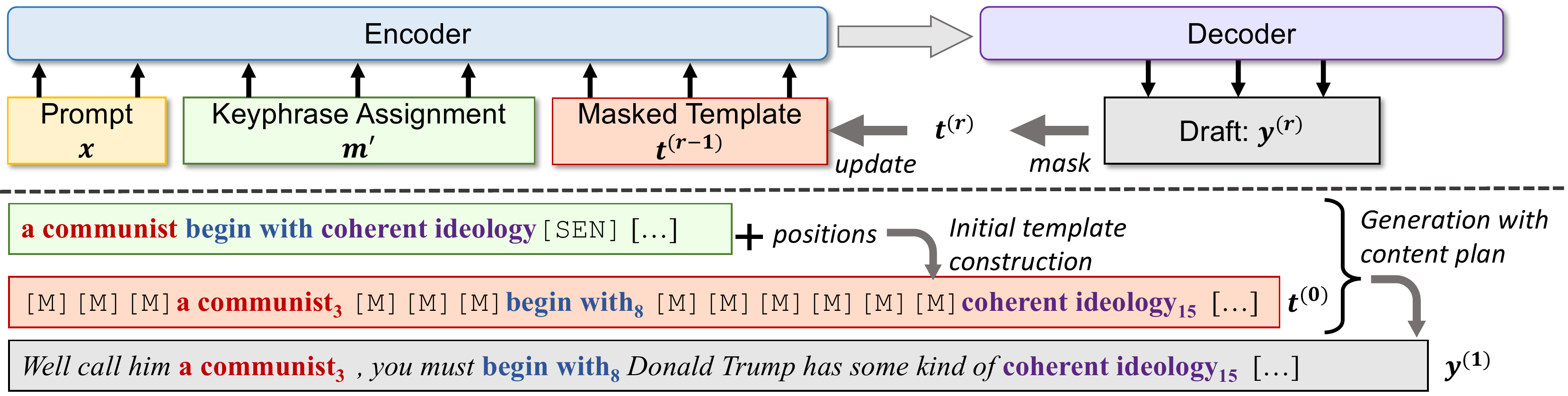}
    \caption{
    Our content-controlled text generation framework, \textsc{PAIR}, which is built on BART. 
    Decoding is executed iteratively. 
    At each iteration, the encoder consumes the input prompt $\bm{x}$, the keyphrase assignments $\bm{m}'$, as well as a partially masked template   
    ($\bm{t}^{(r-1)}$  for the $r$-th iteration, \texttt{[M]} for masks). 
    The autoregressive decoder produces a complete sequence $\bm{y}^{(r)}$, a subset of which is further masked, to serve as the next iteration's template $\bm{t}^{(r)}$.
    }
    \label{fig:model-overview}
\end{figure*}

\subsection{Adding Content Plan with a Template Mask-and-Fill Procedure}
\label{sec:realization}

Given a content planning model, we invoke it to output keyphrase assignments to different sentences ($\bm{m}'$), their corresponding positions $\bm{s}$, along with each sentence's length (based on the prediction of \texttt{[SEN]}). 
We first employ a post-processing step to convert between different tokenizers, and correct erroneous position predictions that violate the assignment ordering or break the consecutivity of the phrase (Appendix~\ref{sec:reprod}). 
We then convert the plan into a {\bf template} $\bm{t}^{(0)}$ as follows: For each sentence, the assigned keyphrases are placed at their predicted positions, and empty slots are filled with \texttt{[MASK]} symbols. 
Figure~\ref{fig:model-overview} illustrates the template construction process and our seq2seq generation model. 
In Appendix~\ref{sec:template-stats}, we show statistics on the constructed templates. 

The input prompt $\bm{x}$, keyphrase assignments $\bm{m}'$, and template $\bm{t}^{(0)}$ are concatenated as the input to the encoder.
The decoder then generates an output $\bm{y}^{(1)}$ according to the model's estimation of $p(\bm{y}^{(1)}|\bm{x}, \bm{m}', \bm{t}^{(0)})$. 
$\bm{y}^{(1)}$ is treated as a draft, to be further refined as described in the next section. 

Our method is substantially different from prior work that uses constrained decoding to enforce words to appear at specific positions~\cite{hokamp-liu-2017-lexically,post-vilar-2018-fast,hu-etal-2019-improved}, which is highly biased by the surrounding few words and suffers from disfluency. 
Since BART is trained to denoise the masked input with contextual understanding, it naturally benefits our method.

\smallskip
\noindent \textbf{Decoding.} 
We employ the nucleus sampling strategy~\cite{holtzman2019curious}, which is shown to yield superior output quality in long text generation. 
In addition to the standard top-$k$ sampling from tokens with the highest probabilities, nucleus sampling further limits possible choices based on a cumulative probability threshold (set to $0.9$ in all experiments below). 
We also require the {\it keyphrases to be generated at or nearby their predicted positions}. 
Concretely, for positions that match any keyphrase token, we force the decoder to copy the keyphrase unless it has already been generated in the previous five tokens. 
We sample three times to choose the one with the lowest perplexity, as estimated by $\text{GPT-2}_{\text{base}}$~\cite{radford2019language}.

\subsection{Iterative Refinement}
\label{sec:refinement}

Outputs generated in a single pass may suffer from incorrectness and incoherence (see Figure~\ref{fig:intro-example}), therefore we propose an iterative refinement procedure to improve the quality.
In each pass, tokens with low generation confidence are masked (Algorithm~\ref{alg:mask-and-refine}).
This is inspired by iterative decoding designed for inference acceleration in {\it non-autoregressive} generation~\cite{lee-etal-2018-deterministic,lawrence-etal-2019-attending}, though their refinement mostly focuses on word substitution and lacks the flexibility for other operations. 
Moreover, our goal is to improve fluency while ensuring the generation of given keyphrases.

At each iteration, the $n$ least confident tokens are replaced with \texttt{[MASK]}. 
Similar as the mask-predict algorithm~\cite{ghazvininejad-etal-2019-mask}, we gradually reduce the number of masks.
In our experiments, each sample is refined for $5$ iterations, with $n$ decaying linearly from $80\%$ of $|\bm{y}^{(r)}|$ to $0$.

\begin{algorithm}[t]
 
\KwData{prompt $\bm{x}$, keyphrase assignments $\bm{m}'$, keyphrase positions $\bm{s}$,
        $R$ refinement iterations, $\rho$ nucleus sampling runs}
\KwResult{final output $\bm{y}^{(R)}$}
 Construct template $\bm{t}^{(0)}$ based on $\bm{m}'$ and $\bm{s}$ \;
 \For{$r=1$ \KwTo $R$}{
    Run encoder over $\bm{x} \oplus \bm{m}' \oplus \bm{t}^{(r-1)}$ \;
    $\mathcal{Y} \leftarrow \varnothing$ \;
    
    \For{$i=1$ \KwTo $\rho$}{
        Run nucleus sampling to generate $\bm{y}_i$ with keyphrase position enforcement\;
        Append $\bm{y}_i$ to $\mathcal{Y}$\;
    }
    $\bm{y}^{(r)} \leftarrow \argmin_{\bm{y}_i \in \mathcal{Y}} \texttt{GPT2-PPL}(\bm{y}_i)$\;
    
    $n \leftarrow |\bm{y}^{(r)}| \times(1 - r/R)$\;
    Mask $n$ tokens with the lowest probabilities to create new template $\bm{t}^{(r)}$\;
 }
 \caption{
 Iteratively refinement via template mask-and-fill.
 The sample with the lowest perplexity (thus with better fluency) is selected for each iteration. 
 }
 \label{alg:mask-and-refine}
\end{algorithm}

\smallskip
\noindent \textbf{Training the Generator.} 
Our training scheme is similar to masked language model pre-training. Given the training corpus $\mathcal{D}=\{(\bm{x}_i, \bm{m}'_i, \bm{y}_i)\}$, we consider two approaches that add noise to the target $\bm{y}_i$ by randomly masking a subset of (1) any tokens, or (2) tokens that are not within the span of any keyphrase. The latter is better aligned with our decoding objective, since keyphrases are never masked. 
We concatenate $\bm{x}_i$, $\bm{m}'_i$, and the corrupted target $\widetilde{\bm{y}_i}$ as input, and fine-tine BART to reconstruct the original $\bm{y}_i$ with a cross-entropy loss.

\section{Experiment Setups}
\label{sec:exp}
\subsection{Tasks and Datasets}
We evaluate our generation and planning models on datasets from three distinct domains for multi-paragraph-level text generation: 
(1) argument generation (\textbf{\textsc{ArgGen}})~\cite{hua-etal-2019-argument-generation}, to produce a counter-argument to refute a given proposition; 
(2) writing opinionated articles (\textbf{\textsc{Opinion}}), e.g., editorials and op-eds, to show idea exchange on a given subject; 
and (3) composing news reports (\textbf{\textsc{News}}) to describe events. 
The three domains are selected with diverse levels of subjectivity and various communicative goals (persuading vs. informing), with statistics shown in Table~\ref{tab:data-stats}. 

\begin{table}[th]
\fontsize{10}{11}\selectfont
    \centering
    \setlength{\tabcolsep}{1.0pt}
    \begin{tabular}{l lllll}
     \toprule
        & {\bf\# Sample} & $\vert${\bf Prompt}$\vert$ &  $\vert${\bf Target}$\vert$ &  {\bf \# KP} & {\bf KP Cov.} \\
        \midrule
        \textsc{ArgGen} & 56,504 & 19.4 &   116.6 & 20.6 & 30.5\% \\
        \textsc{Opinion} & 104,610 & 6.1 &  205.6 &  19.0 & 26.0\% \\
        \textsc{News} & 239,959 & 7.0 &  282.7 & 30.3 & 32.6\% \\
        \bottomrule
    \end{tabular}
    \caption{
    Statistics of the three datasets. We report average lengths of the prompt and the target generation, number of unique keyphrases (\# KP) used in the input, and the percentage of content words in target covered by the keyphrases (KP Cov.).
    }
    \label{tab:data-stats}
\end{table}

\noindent \textbf{Task 1: Argument Generation.} 
We first evaluate our models on persuasive argument generation, based on a dataset collected from Reddit \texttt{r/ChangeMyView} (CMV) in our prior work~\cite{hua-etal-2019-argument-generation}. This dataset contains pairs of original post (OP) statement on a controversial issue about politics and filtered high-quality counter-arguments, covering $14,833$ threads from $2013$ to $2018$.
We use the OP title, which contains a proposition (e.g. \textit{the minimum wage should be abolished}), to form the input prompt $\bm{x}$. 
In our prior work, only the first paragraphs of high-quality counter-arguments are used for generation. Here we consider generating the full post, which is significantly longer.
Keyphrases are identified as noun phrases and verb phrases that contain at least one topic signature word~\cite{C00-1072}, which is determined by a log-likelihood ratio test that indicates word salience. 
Following our prior work, we expand the set of topic signatures with their synonyms, hyponyms, hypernyms, and antonyms according to WordNet~\cite{miller-1994-wordnet}. The keyphrases longer than $10$ tokens are further discarded.

\begin{table*}[t]
\centering
\fontsize{10}{11}\selectfont
 \setlength{\tabcolsep}{1.2mm}
\centering
    \begin{tabular}{l llll l llll l llll}
    \toprule
        & \multicolumn{4}{c}{\textsc{ArgGen}} & \phantom{} & \multicolumn{4}{c}{\textsc{Opinion}} & \phantom{} &
        \multicolumn{4}{c}{\textsc{News}}
        \\
        
         & \textbf{B-4} & \textbf{R-L} & \textbf{MTR} & \textbf{Len.} &
         \phantom{}  &
         \textbf{B-4} & \textbf{R-L} & \textbf{MTR} &  \textbf{Len.} &
         \phantom{} & 
         \textbf{B-4} &  \textbf{R-L} & \textbf{MTR} &  \textbf{Len.}
         \\
        
        \midrule
        
        \textsc{Seq2seq} & 0.76 & 13.80 & 9.36 & 97 & 
        \phantom{} &
        1.42 & 15.97 & 10.97 & 156 & 
        \phantom{} &
        1.11 & 15.60 & 10.10 & 242 \\
        
        \textsc{KPSeq2seq} & 6.78 & 19.43 & 15.98 & 97  & 
        \phantom{} &
        11.38 & 22.75 & 18.38 & 164  &
        \phantom{}
        & 11.61 & 21.05 & 18.61 & 286 \\

        \hdashline
        
       {\textsc{PAIR}$_\text{light}$}   & 26.38 &  47.97 & 31.64 & 119 & 
        \phantom{} &
        16.27 & 33.30 & 24.32 &210 &
        \phantom{} &
        28.03 & 43.39 & 27.70 &  272  \\
        
      {\textsc{PAIR}$_\text{light}$} w/o refine & 25.17 &  46.84 & 31.31 & 120 & 
        \phantom{} &
        15.45 & 32.35 & 24.11 & 214 & 
        \phantom{} &
        27.32 &  43.08 & 27.35 & 278 \\

       {\textsc{PAIR}$_\text{full}$} & {\bf 36.09} &  {\bf 56.86} & {\bf 33.30}  & 102 & 
        \phantom{} &
        {\bf 23.12} &  {\bf 40.53} & {\bf 24.73} &  167 &
        \phantom{}
        & {\bf 34.37}  & {\bf 51.10} & {\bf 29.50} & 259 \\

        {\textsc{PAIR}$_\text{full}$} w/o refine & 34.09 &  55.42 & 32.74 & 101 &
        \phantom{} &
        22.17 &  39.71 & 24.65 & 169 &
        \phantom{} &
        33.48 &  50.27 & 29.26 & 260 \\
     \bottomrule
    \end{tabular}
 
    \caption{
    Key results on argument generation, opinion article writing, and news report generation. BLEU-4 (B-4), ROUGE-L (R-L), METEOR (MTR), and average output lengths are reported (for references, the lengths are 100, 166, and 250, respectively).
    {\textsc{PAIR}}$_\text{light}$, using keyphrase assignments only, consistently outperforms baselines; adding keyphrase positions, {\textsc{PAIR}}$_\text{full}$ further boosts scores. Improvements by our models over baselines are all significant ($p<0.0001$, approximate randomization test). 
    Iterative refinement helps on both setups. 
  }
  \label{tab:main-results}
\end{table*}

\smallskip
\noindent \textbf{Task 2: Opinion Article Generation.}
We collect opinion articles from the New York Times (NYT) corpus~\cite{sandhaus2008new}. An article is selected if its \texttt{taxonomies} label has a prefix of \textit{Top/Opinion}. We eliminate articles with an empty headline or less than three sentences. 
Keyphrases are extracted in a similar manner as done in argument generation. Samples without any keyphrase are removed. 
The article headline is treated as the input, and our target is to construct the full article. 
Table~\ref{tab:data-stats} shows that opinion samples have shorter input than arguments, and the keyphrase set also covers fewer content words in the target outputs, requiring the model to generalize well to capture the unseen tokens.

\smallskip
\noindent \textbf{Task 3: News Report Generation.} 
Similarly, we collect and process news reports from NYT, filtering by taxonomy labels starting with ``\textit{Top/News}", removing articles that have no content word overlap with the headline, and ones with \texttt{material-types} labeled as one of ``\textit{statistics}'', ``\textit{list}'', ``\textit{correction}'', ``\textit{biography}'', or ``\textit{review}.'' 
News reports describe events and facts, and in this domain we aim to study and emphasize the importance of faithfully reflecting content plans during generation and refinement. 

\smallskip
\noindent \textbf{Data Split and Preprocessing.}
For argument generation, we split the data into $75\%$, $12.5\%$, and $12.5\%$ for training, validation, and test sets. To avoid test set contamination, the split is conducted on thread level.
For opinion and news generation, we reserve the most recent $5$k articles for testing, another $5$k for validation, and the rest ($23$k for news and $10$k for opinion) are used for training. 
We apply the BPE tokenization~\cite{sennrich-etal-2016-neural} for the generation model as BART does, and use WordPiece~\cite{wu2016google} for BERT-based planner. To fit the data into our GPUs, we truncate the target size to $140$ tokens for argument, sizes of $243$ and $335$ are applied for opinion and news, for both training and inference.

\subsection{Implementation Details}

Our code is written in PyTorch~\cite{paszke2019pytorch}. 
For fine-tuning, we adopt the standard linear warmup and inverse square root decaying scheme for learning rates, with a maximum value of $5\times 10^{-5}$. Adam~\cite{kingma2014adam} is used as the optimizer, with a batch size of $10$ for refinement and $20$ for content planning, and a maximum gradient clipped at $1.0$. 
All hyperparameters are tuned on validation set, with early stopping used to avoid overfitting.
More details are in Appendix~\ref{sec:reprod}.

\subsection{Baselines and Comparisons}

We consider two baselines, both are fine-tuned from BART as in our models: (1) \textbf{\textsc{Seq2seq}} directly generates the target from the prompt; (2) \textbf{\textsc{KPSeq2seq}} encodes the concatenation of the prompt and the \textit{unordered} keyphrase set. 
To study if using only sentence-level keyphrase assignments helps, we include a model variant (\textbf{\textsc{PAIR}$_{\text{light}}$}) by removing keyphrase position information ($\bm{s}$) from the input of our generator and using an initial template with all \texttt{[MASK]} symbols.
Our model with full plans is denoted as \textbf{\textsc{PAIR}$_{\text{full}}$}. 
We first report generation results using {\it ground-truth content plans} constructed from human-written text, and also show the end-to-end results with {\it predicted content plans} by our planner.

\section{Results}
\label{sec:results}
\subsection{Automatic Evaluation}
\label{sec:auto-eval}
We report scores with BLEU~\cite{papineni-etal-2002-bleu}, which is based on n-gram precision (up to 4-grams); ROUGE-L~\cite{lin-2004-rouge}, measuring recall of the longest common subsequences; and METEOR~\cite{lavie-agarwal-2007-meteor}, which accounts for paraphrase. 
For our models PAIR$_\text{full}$ and PAIR$_\text{light}$, we evaluate both the first draft and the final output after refinement. 
Table~\ref{tab:main-results} lists the results when ground-truth content plans are applied.

First, \textit{our content-controlled generation model with planning consistently outperforms comparisons and other model variants on all datasets}, with or without iterative refinement. Among our model variants, PAIR$_\text{full}$ that has access to full content plans obtains significantly better scores than PAIR$_\text{light}$ that only includes keyphrase assignments but not their positions. Lengths of PAIR$_\text{full}$'s outputs are also closer to those of human references. Both imply the benefit of keyphrase positioning.

\begin{figure}[t]
\hspace{-2mm}
\includegraphics[width=78mm]{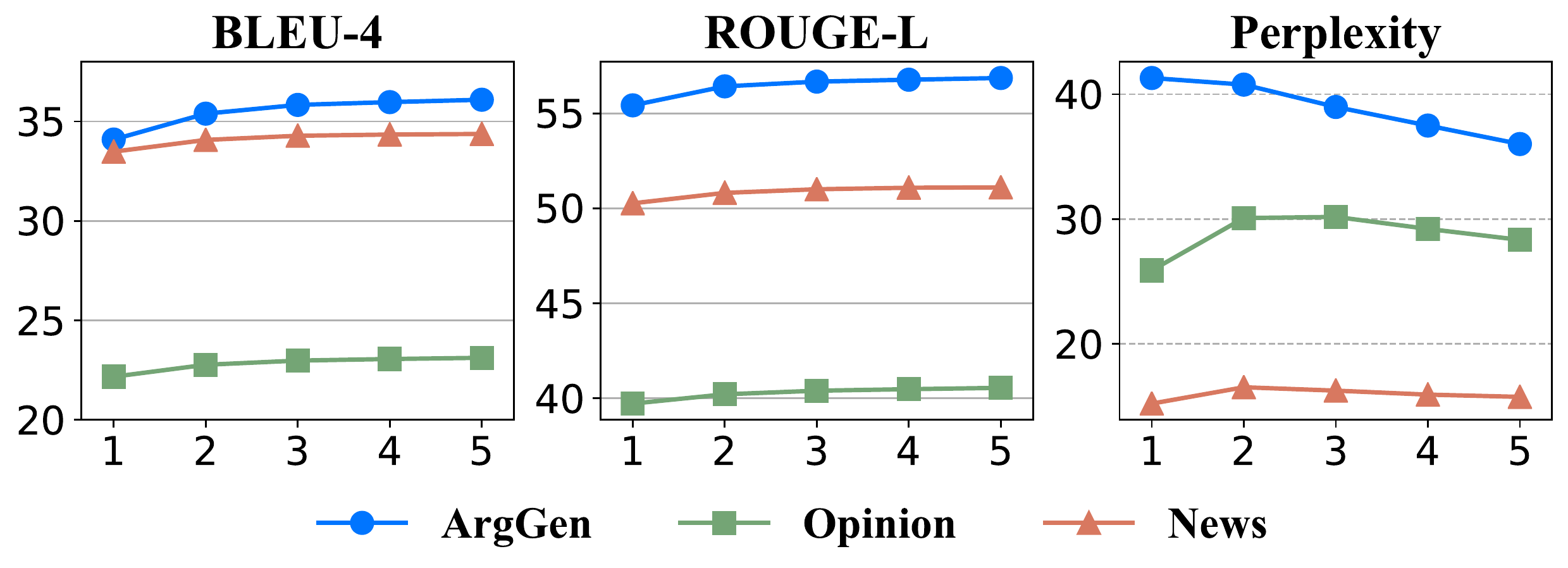}
\caption{
Results on iterative refinement with five iterations. Both BLEU and ROUGE-L scores steadily increase, with perplexity lowers in later iterations. 
}
\label{fig:refine_change}
\end{figure}

Table~\ref{tab:main-results} also shows that \textit{the iterative refinement strategy can steadily boost performance} on both of our setups.
By inspecting the performance of refinement in different iterations (Figure~\ref{fig:refine_change}), we observe that both BLEU and ROUGE-L scores gradually increase while perplexity lowers as the refinement progresses.
This indicates that iterative post-editing improves both content and fluency. 

\begin{figure}[t]
\hspace{-1mm}
\includegraphics[width=79mm]{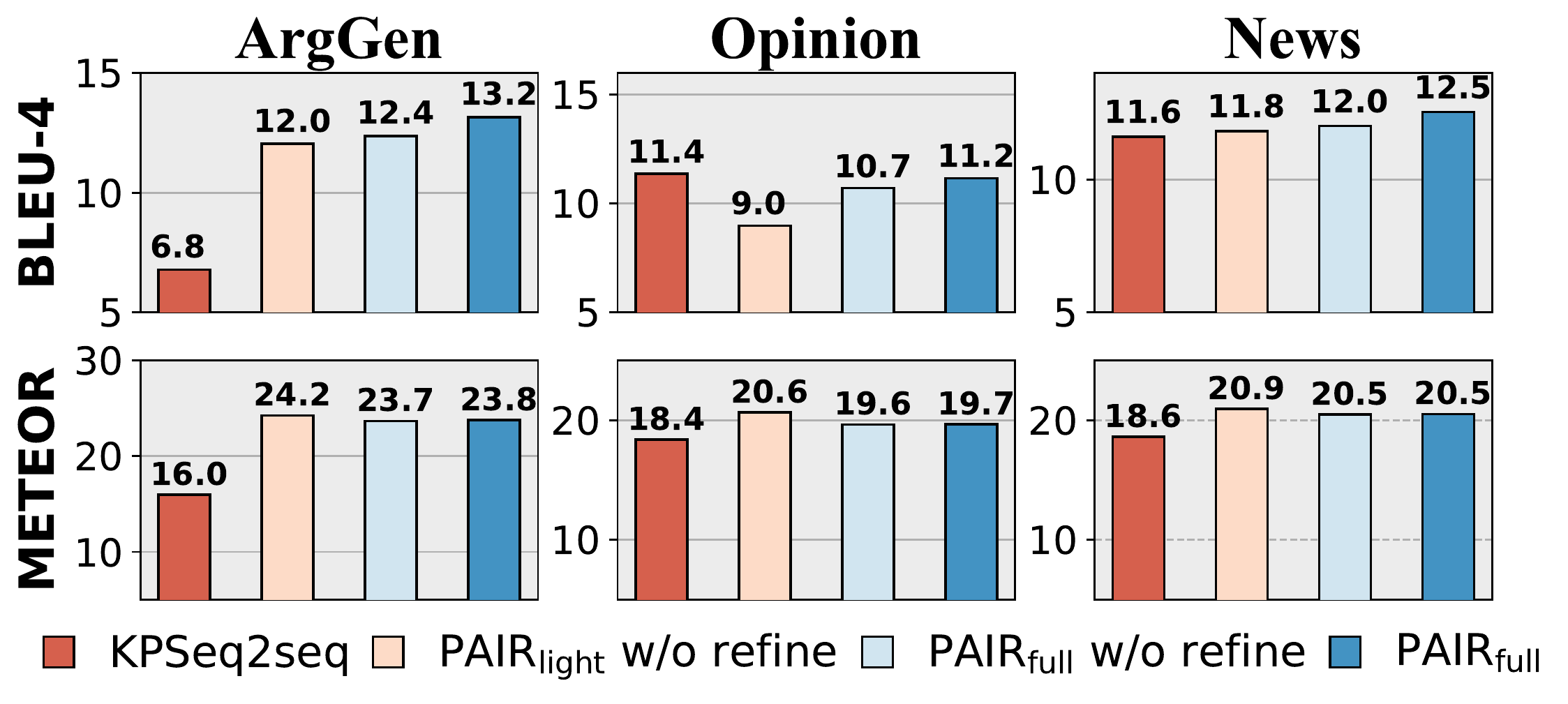}
\caption{
End-to-end generation results with automatically predicted content plans.
Our models outperform \textsc{KPSeq2seq} in both metrics, except for BLEU-4 on opinion articles where results are comparable. 
}
\label{fig:system-results}
\end{figure}

\smallskip
\noindent \textbf{Results with Predicted Content Plans.} 
We further report results by using content plans predicted by our BERT-based planner. 
Figure~\ref{fig:system-results} compares PAIR$_\text{full}$ and PAIR$_\text{light}$ with \textsc{KPSeq2seq}. 
Our models yield better METEOR scores on all three domains.
That said, the improvement from predicted plans is not as pronounced as that from ground-truth plans. Upon inspection, we find that our planner often falls short of accurately positioning the given keyphrases, leading to degraded generation performance. This points to a potential direction for future work where better positioning model should be developed.

\subsection{Human Evaluation}
\label{sec:human}
We hire four proficient English speakers\footnote{They are all US-based college students. Each of them is paid \$15 hourly for the task.} to rate three aspects of the generated arguments on a scale of $1$ (worst) to $5$ (best): \textbf{fluency}, \textbf{coherence}---if the information organization is natural and logical, and \textbf{relevance}---if the topic is related to the prompt and whether the stance is correct. 
$50$ samples are randomly selected, with system outputs by \textsc{KPSeq2seq}, PAIR$_\text{full}$ and PAIR$_\text{light}$ shown to human judges in random order. 
The evaluation guideline is in the supplementary material.

\begin{table}[t]
\fontsize{10}{12}\selectfont
 \setlength{\tabcolsep}{1mm}
  \centering
    \begin{tabular}{llll}
        \toprule
        \textsc{ArgGen} & {\bf Fluency} & {\bf Coherence} & {\bf Relevance}\\
        \midrule
        \textsc{KPSeq2seq} & 4.63  & 3.28 & 2.79 \\
        \hdashline
        {\textsc{PAIR}$_\text{light}$} & {\bf 4.75} & {\bf 3.97}$^{\ast}$ & {\bf 3.85}$^{\ast}$ \\
        {\textsc{PAIR}$_\text{full}$} & 4.46 & 3.76$^{\ast}$ & 3.79$^{\ast}$  \\
        \bottomrule
    \end{tabular}
    
    \caption{
    Human evaluation for argument generation on fluency, coherence, and relevance, with 5 as the best.  
    The Krippendorff's $\alpha$ are 0.28, 0.30, and 0.37, respectively. 
    Our model outputs are significantly more coherent and relevant than \textsc{KPSeq2seq} ($^{\ast}$: $p<0.0001$), with comparable fluency.
  }
  \label{tab:human-eval-arggen}
\end{table}

\definecolor{byzantine}{rgb}{0.74, 0.2, 0.64}
\definecolor{darkraspberry}{rgb}{0.53, 0.15, 0.34}
\definecolor{rust}{rgb}{0.72, 0.25, 0.05}
\definecolor{princetonorange}{rgb}{1.0, 0.56, 0.0}
\definecolor{amber}{rgb}{1.0, 0.75, 0.0}
\definecolor{ao}{rgb}{0.0, 0.5, 0.0}
\definecolor{bondiblue}{rgb}{0.0, 0.58, 0.71}
\definecolor{blush}{rgb}{0.87, 0.36, 0.51}
\definecolor{chartreuse}{rgb}{0.5, 1.0, 0.0}
\definecolor{deeplilac}{rgb}{0.6, 0.33, 0.73}
\definecolor{dogwoodrose}{rgb}{0.84, 0.09, 0.41}
\definecolor{munsell}{rgb}{0.0, 0.66, 0.47}
\definecolor{jade}{rgb}{0.0, 0.66, 0.42}
\begin{table}[t]
\hspace{-2mm}
\fontsize{9}{11}\selectfont
\setlength{\tabcolsep}{1.0mm}
    \begin{tabular}{p{76mm}}

    \toprule
    \textbf{Prompt} (News): \textit{4 Arrested in Theft of Baby Jesus Figurines} \\
    \midrule
    \textbf{PAIR$_\text{full}$}: {\color{bondiblue}{\bf Four New Jersey teenagers arrested yesterday}} were accused of stealing more than {\color{bondiblue}{\bf 25 plastic baby Jesus figurines}} from a church before they {\color{bondiblue}{\bf burn in a bonfire}}, {\color{bondiblue}{\bf the police}} said. 
    {\color{dogwoodrose}{\bf The police}} in {\color{dogwoodrose}{\bf Sayreville}}, N.J., {\color{dogwoodrose}{\bf arrested Michael Payne, 18}}, and T.J. Jones, 18, of Union City; {\color{dogwoodrose}{\bf Nicholas M. Hess}}, 18, of {\color{dogwoodrose}{\bf Matawan}}; and Jason L. O'Neill, 18, of Port New York, N.J., and charged them with burglary. Their vandals removed more than 100 figurines of Jesus from {\color{deeplilac}{\bf a cemetery}} outside {\color{deeplilac}{\bf St. Stanislaus Kostka Church}} in Sayreville, the police said. Detectives said the four had taken their {\color{jade}{\bf flashlights}} to {\color{jade}{\bf the cemetery}} and {\color{jade}{\bf jotted down the license plates}} of several cars in {\color{jade}{\bf the vicinity}} [\ldots] \\
    \midrule
    \textbf{Prompt} (Opinion): \textit{Drowning in a Sea of College Debt} \\
    \midrule
    \textbf{PAIR$_\text{full}$}: { Once again, the appalling problem of {\color{blue}{\bf increasing debt}} in {\color{blue}{\bf financing higher education}} that has taken place is more {\color{blue}{\bf intolerable}}, particularly because of the arrogance of our colleges and universities. The burden of {\color{red}{\bf saddling college}} students with debt in the middle of their {\color{red}{\bf teenage years}}, when they were in debt, is essential for {\color{red}{\bf a good education}}. Our educational system is designed to allow kids to develop the skills necessary, but it does not {\color{darkraspberry}{\bf create optimal conditions}} for mature students who know they will not be able} [\ldots]\\
    \bottomrule
    \end{tabular}
\caption{
Sample outputs in the news and opinion domain. Keyphrases assigned to different sentences are in boldface and color-coded. 
}
\label{tab:sample-output}
\end{table}

Table~\ref{tab:human-eval-arggen} shows that both of our models
are rated with better coherence and relevance than \textsc{KPSeq2seq} which uses the same but unordered keyphrases as input. Interestingly, outputs by PAIR$_\text{light}$ are regarded as more fluent and coherent, though the difference is not significant. However, discourse analysis in \S~\ref{sec:discussion} reveals that clauses produced by PAIR$_\text{light}$ are more locally related, compared to PAIR$_\text{full}$, which can be perceived as easier to read. 
In addition to the sample argument in Figure~\ref{fig:intro-example}, Table~\ref{tab:sample-output} shows PAIR$_\text{full}$'s output in the news and opinion domains. 
More samples by different systems are in the supplementary material. 

\noindent \textbf{Effect of Refinement and Keyphrase Enforcement.} 
We further ask {\it whether human judges prefer the refined text} and {\it whether enforcing keyphrases to be generated yields noticeable content improvement}. 
In a second study, we present the same $50$ prompts from the previous evaluation on argument generation, and an additional $50$ samples for opinion article writing to the same group of human judge. 
For each sample, PAIR$_\text{full}$'s outputs with and without refinement are shown in random order. Judges indicate their preference based on the overall quality. 
The same procedure is conducted to compare with a version where we do not enforce keyphrases to be copied at their predicted positions during decoding. 
Table~\ref{tab:ab_testing} demonstrates that the refined text is preferred in more than half of the cases, for both domains.
Enforcing keyphrase generation based on their positions is also more favorable than not enforcing such constraint. 

\begin{table}[h]
\fontsize{10}{11}\selectfont
 \setlength{\tabcolsep}{.8mm}
  \centering
\begin{tabular}{lcccc}

& \multicolumn{1}{c}{\textsc{PAIR}$_\text{full}$} & \multicolumn{1}{c|}{{\it w/o refine}} & \multicolumn{1}{c}{\textsc{PAIR}$_\text{full}$} & \multicolumn{1}{c}{{\it w/o enforce}} \\
\textsc{ArgGen} &  {\bf 52.7\%} & \multicolumn{1}{c|}{33.3\%} & {\bf 45.3\%} & 40.0\%  \\
\textsc{Opinion} & {\bf 52.7\%} & \multicolumn{1}{c|}{30.7\%} & {\bf 50.0\%}  & 29.3\% \\
\end{tabular}
    \caption{
    Percentages of samples preferred by human judges before and after refinement [Left]; with and without enforcing keyphrases to appear at the predicted positions [Right]. Ties are omitted.
    }
    \label{tab:ab_testing}
\end{table}

\noindent \textbf{What is updated during iterative refinement?} 
Since refinement yields better text, we compare generations before and after the refinement. 
First, we find that masks are regularly put on ``functional" words and phrases. For example, stopwords and punctuation along with their bigrams are often swapped out, with new words filled in to improve fluency.  Moreover, about $85\%$ of the refinement operations result in new content being generated.
This includes changing prepositions and paraphrasing, e.g., replacing ``\textit{a research fellow}'' with ``\textit{a graduate student}.'' 
On both news and opinion domains, numerical and temporal expressions are often incorrectly substituted, suggesting that better fact control needs to be designed to maintain factuality.

\section{Further Discussions on Discourse}
\label{sec:discussion}
\renewcommand{\UrlFont}{\small}

Prior work's evaluation mainly focuses on fluency and content relevance, and largely ignores the discourse structure exposed by the generated text.
However, unnatural discourse and lack of focus are indeed perceived as major problems of long-form neural generations, as identified by human experts.\footnote{\url{https://www.economist.com/open-future/2019/10/01/how-to-respond-to-climate-change-if-you-are-an-algorithm}} 
Here, we aim to investigate whether content-controlled generation with ground-truth content plans resembles human-written text by studying discourse phenomena.

\begin{figure}[t]
    \hspace{-2mm}
    \includegraphics[width=78mm]{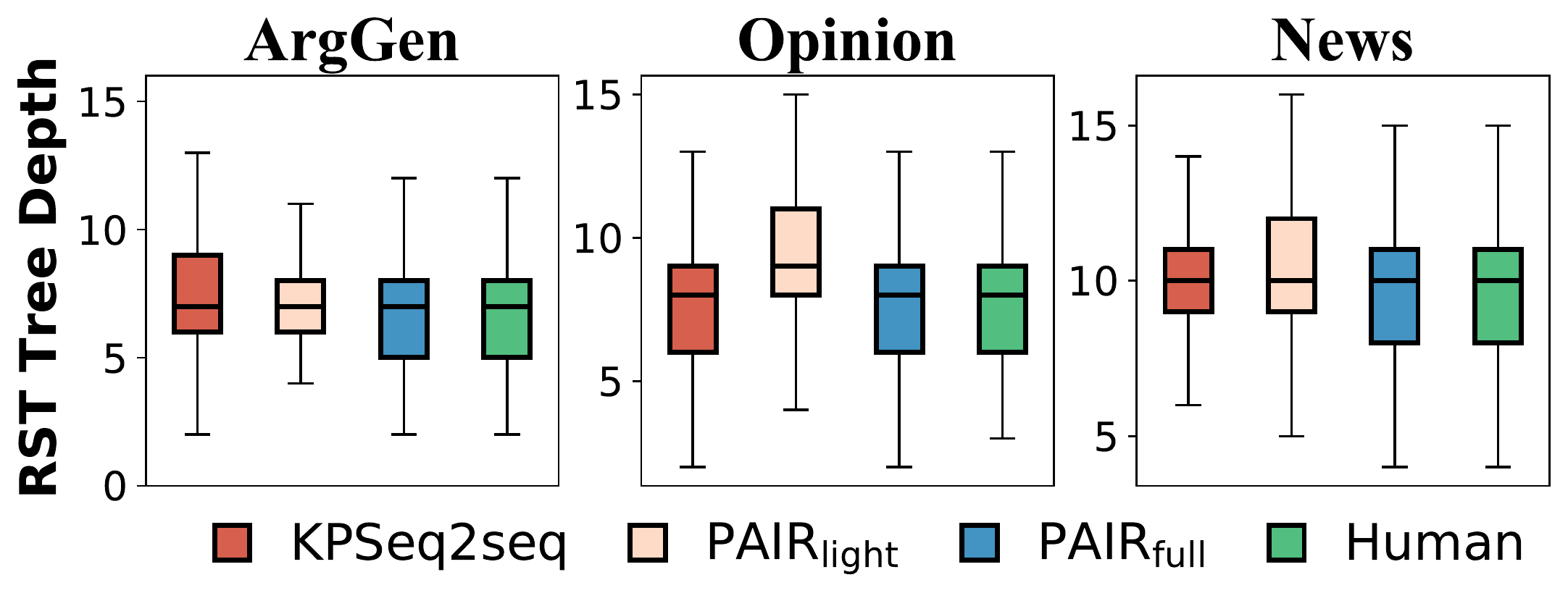}
    \caption{
    Distributions of RST tree depth.
    PAIR$_\text{full}$ better resembles the patterns in human-written texts.
    }
    \label{fig:rst-depth}
\end{figure}

\smallskip
\noindent \textbf{Are PAIR generations similar to human-written text in discourse structure?} 
We utilize DPLP~\cite{ji-eisenstein-2014-representation}, an off-the-shelf Rhetorical Structure Theory (RST) discourse parser. DPLP converts a given text into a binary tree, with elementary discourse units (EDUs, usually clauses) as nucleus and satellite nodes. For instance, a relation \texttt{NS-elaboration} indicates the second node as a satellite (\texttt{S}) elaborating on the first nucleus (\texttt{N}) node. DPLP achieves F1 scores of $81.6$ for EDU detection and $71.0$ for relation prediction on news articles from the annotated RST Discourse Treebank~\cite{carlson-etal-2001-building}. We run this trained model on our data for both human references and model generations. 

First, we analyze the {\it depth of RST parse trees}, which exhibits whether the text is more locally or globally connected. For all trees, we truncate at a maximum number of EDUs based on the $90$ percentile of EDU count for human references. 
Distributions of tree depth are displayed in Figure~\ref{fig:rst-depth}. As can be seen, generations by PAIR$_\text{full}$ show similar patterns to human-written arguments and articles. We also find that trees by PAIR$_\text{light}$ tend to have a more ``linear" structure, highlighting the dominance of local relations between adjacent EDUs, compared with PAIR$_\text{full}$ which uses knowledge of keyphrases positions.
This implies that content positioning helps with structure at a more global level. 
We further look into the {\it ratios of \texttt{NS}, \texttt{NN}, \texttt{SN} relations}, and observe that most model outputs have similar trends as human-written texts, except for \textsc{KPSeq2seq} which has more \texttt{SN} relations, e.g., it produces twice as many \texttt{SN}s than others on arguments. 

\begin{figure*}[h]
\centering
\includegraphics[width=150mm]{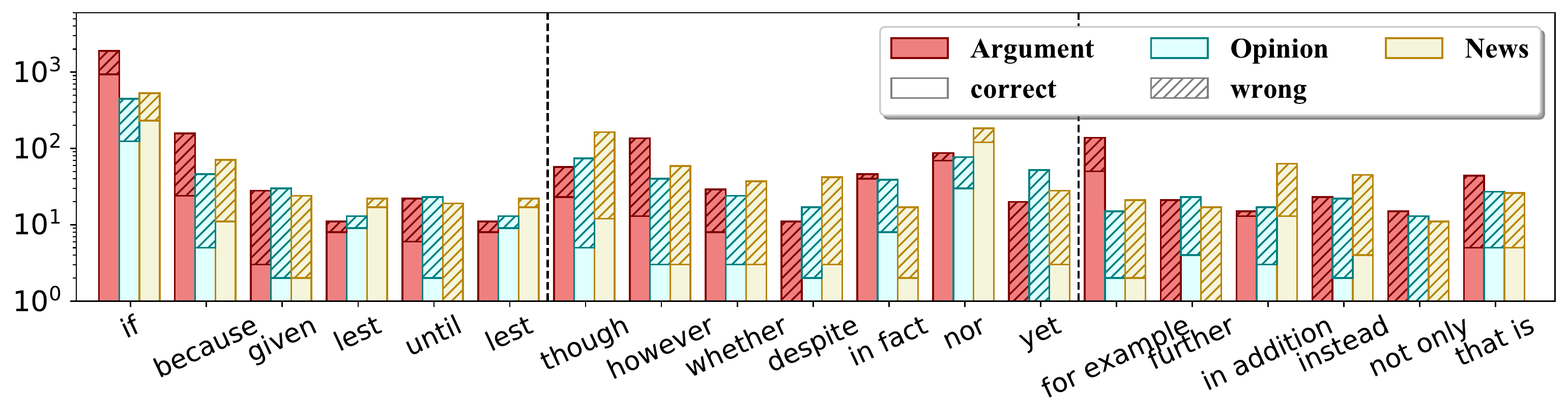}
\caption{
Discourse markers that are correctly and incorrectly (shaded) generated by PAIR$_\text{full}$, compared to aligned sentences in human references. 
Discourse markers are grouped (from left to right) into senses of \textsc{Contingency} (higher marker generation accuracy observed), \textsc{Comparison}, and \textsc{Expansion}. 
$y$-axis: \# of generated sentences with the corresponding marker. 
}
\label{fig:disc-marker}
\end{figure*}

\smallskip
\noindent \textbf{Can PAIR correctly generate discourse markers?} 
Since discourse markers are crucial for coherence~\cite{grote-stede-1998-discourse,callaway-2003-integrating} and have received dedicated research efforts in rule-based systems~\cite{reed-etal-2018-neural,balakrishnan-etal-2019-constrained}, we examine if PAIR$_\text{full}$ can properly generate them. 
For each sample, we construct sentence pairs based on content word overlaps between system generation and human reference. 
We manually select a set of unambiguous discourse markers from Appendix A of the Penn Discourse Treebank manual~\cite{prasad2008penn}. 
When a marker is present in the first three words in a reference sentence, we check if the corresponding system output does the same.

Figure~\ref{fig:disc-marker} displays the numbers of generated sentences with markers produced as the same in human references (correct) or not (wrong). 
The markers are grouped into three senses: \textsc{Contingency}, \textsc{Comparison}, and \textsc{Expansion}. 
The charts indicates that PAIR$_\text{full}$ does better at reproducing markers for \textsc{Contingency}, followed by \textsc{Comparison} and \textsc{Expansion}. 
Manual inspections show that certain missed cases are in fact plausible replacements, such as using \texttt{at the same time} for \texttt{in addition}, or \texttt{also} for \texttt{further},  while in other cases the markers tend to be omitted. 
Overall, we believe that content control alone is still insufficient to capture discourse relations, motivating future work on discourse planning.

\section{Ethics Statement}
\label{sec:ethical}
We recognize that the proposed system can generate fabricated and inaccurate information due to the systematic biases introduced during model pre-training based on web corpora. We urge the users to cautiously examine the ethical implications of the generated output in real world applications.

\section{Conclusion}
\label{sec:conclusion}
We present a novel content-controlled generation framework that adds content planning to large pre-trained Transformers without modifying model architecture. A BERT-based planning model is first designed to assign and position keyphrases into different sentences. 
We then investigate an iterative refinement algorithm that works with the sequence-to-sequence models to improve generation quality with flexible editing.
Both automatic evaluation and human judgments show that our model with planning and refinement enhances the relevance and coherence of the generated content. 

\section*{Acknowledgements}
This research is supported in part by National Science Foundation through Grant IIS-1813341 and Nvidia GPU gifts. We thank three anonymous reviewers for their constructive suggestions on many aspects of this work. 

\bibliographystyle{acl_natbib}
\bibliography{related}

\appendix
\section{Reproducibility}
\label{sec:reprod}
\noindent\textbf{Computing Infrastructure.}
Our model is built upon the PyTorch \texttt{transformers-2.6.0} library by \citet{Wolf2019HuggingFacesTS}, with 
\texttt{Pytorch-Lightning-0.7.3}~\cite{falcon2019pytorch} for training routines.
To improve training efficiency, we adopt mixed-precision floating point (FP16) computation using the \texttt{O2} option of NVIDIA \texttt{apex}\footnote{\url{https://github.com/NVIDIA/apex}}.
For both training and decoding, we utilize the Titan RTX GPU card with $24$ GB memory.

\smallskip
\noindent\textbf{Model Sizes.}
Our generation model has the same architecture as BART~\cite{lewis-etal-2020-bart} with $406$M parameters. The content planner is built on top of BERT$_\text{base}$, which has $110$M parameters.

\smallskip
\noindent\textbf{Running Time.}
Training the generation model takes $2.5$ hours for argument, $5$ hours for opinion, and $24$ hours for news. 
The content planning model converges in $2.5$-$4$ hours for three domains.

\smallskip
\noindent\textbf{Decoding Settings.}
At inference time, we set $k=50$, temperature=$1.0$, and $p=0.9$ for nucleus sampling. The relatively large $k$ value is determined based on a pilot study, where we find that the refinement lacks diversity if $k$ is set to small values. 
Moreover, since the Transformer states need to be cached during autoregressive decoding and we perform three complete nucleus sampling runs in each refinement iteration, the GPU memory consumption is substantially increased. We therefore limit the maximum generation steps to $140$ for argument, $243$ and $335$ for opinion and news.

\smallskip
\noindent\textbf{Auto-Correction for Content Plan.}
When the content plan is predicted by the planner, the following post-processing steps are employed prior to the masked template construction: (1) For a predicted keyphrase, its token positions are adjusted to a consecutive segment, so that the phrase is kept intact in the template. (2) If the predicted positions are not monotonic to the assignment ordering, they will be rearranged. For instance, if the assignment contains \texttt{KP}$_1$ $\triangleright$ \texttt{KP}$_2$, but position of \texttt{KP}$_2$ is not strictly larger than that of \texttt{KP}$_1$, we instead place \texttt{KP}$_2$ immediately after \texttt{KP}$_1$ in the template. (3) Finally, since the planner and generator have different subword vocabularies, it is necessary to detokenize the predicted keyphrase assignment, and re-tokenize with the BPE vocabulary of the generator.

\section{Template Construction Statistics}
\label{sec:template-stats}
We characterize the content planning results in Table~\ref{tab:template}.
Specifically, we show the statistics on the automatically created templates based on the planner's output. As we can see, our system predicted templates approach human reference in terms of length, per sentence keyphrase count, and the average keyphrase spacing. 
Sentence segmentation occurs more often in our templates than the reference text, likely due to the frequent generation of \texttt{[SEN]} tokens.

\begin{table}[t]
\centering
\fontsize{10}{11}\selectfont
\setlength{\tabcolsep}{0.7mm}
\begin{tabular}{lllllll}
\toprule
& \multicolumn{2}{c}{\textsc{ArgGen}} & \multicolumn{2}{c}{\textsc{Opinion}} & \multicolumn{2}{c}{\textsc{News}} \\
& \textit{sys} & \textit{ref} & \textit{sys} & \textit{ref} & \textit{sys} & \textit{ref} \\
\midrule
\# tokens & 133.3 & 130.2 & 228.5 & 246.3 & 424.5 & 435.5 \\
\# sentences & 8.6 & 5.6 & 11.1 & 8.2 & 19.2 & 13.5 \\
\# KP per sent.& 2.96 & 3.77 & 2.22 & 2.49 & 3.40 & 3.24 \\
KP distance & 2.61 & 2.95 & 5.70 & 6.02 & 3.76 & 5.08 \\ 
\bottomrule
\end{tabular}
\caption{Statistics on generated templates by our content planner. Tokens are measured in units of WordPiece~\cite{sennrich-etal-2016-neural}. KP distance denotes the average number of tokens between two keyphrases that are in the same sentence. Both system output (\textit{sys}) and human reference (\textit{ref}) are reported.}
\label{tab:template}
\end{table}

\section{Human Evaluation}
\label{sec:human-eval-guideline}

As described in \S~5.2 of the paper, we carry out two human evaluation studies.
In the first study, the goal is to assess the output quality of three aspects. The detailed evaluation guideline and examples are listed in Figure~\ref{tab:human-eval-1} and Figure~\ref{tab:human-eval-2}.

\begin{figure*}[h]
    \centering\fontsize{9}{11}\selectfont
    \begin{tabular}{|p{135mm}|}
         \hline
         This study aims to evaluate three text generation systems for counter-argument generation resembling the reddit ChangeMyView style (CMV). In total, 53 sets of samples will be presented. Each entry starts with a statement that has a stance over a certain topic. Machine generated responses will be listed under the title in random orders.   

         Please first read the title and each of the three system outputs. Then rate each output over the following aspects on a likert scale (1-worst, 5-best). At the end of each entry, please also indicate the overall ranking of the four systems:

        {\fontsize{9}{11}\selectfont
        \begin{itemize}[leftmargin=3mm]
            \item {\bf Fluency:} whether the output is free of grammar errors and easy to read
            {\fontsize{9}{11}\selectfont
            \begin{itemize}[leftmargin=3mm]
                \item 1. the output contains multiple major grammar errors that significantly reduce readability, e.g., \textit{``It suggesting looks you that eu perhaps a higher tax rate.''}.
                
                \item 3. the output contains at most one major grammar error, or up to three minor grammar errors, e.g., \textit{``Gender make complete senses, but not so with you. All sex is is the difference between masculinity and femininity.''}
                
                \item 5. the output is fluent and free of any grammar errors, e.g., \textit{``Perhaps the name "Aesop" is a reference to the religious philosophy of this sect, which channels Ramtha, as a spokesman for the Catholic Church.''}
                
            \end{itemize}}
            \item {\bf Coherence:}  whether the information transition is natural and well-structured
            \begin{itemize}[leftmargin=3mm]
                \item 1. the output either has obvious self-contradiction or at least two major incoherent sections, e.g., \textit{``The EU is the way forward. Its not that different from other empires that didn’t work out at the end.''}

                \item 3. the output contains at most one major incoherent section or up to three minor incoherent problems, e.g., \textit{``It may be that you have a ticket to die and you need to take it. That’s why we are soldiers.''}

                \item 5. the information transition is natural and the overall message is clearly delivered, e.g., \textit{``The primary advantage a EU military has is the authority of the EU institutions, without which individual states cannot have coalitions on this level.''}

            \end{itemize}
            \item {\bf Relevance:} whether the response is on topic and has a clear opposing stance
            \newline \quad\quad Topic: \textit{We shouldn't raise the minimum wage to \$15 an hour.}
            \begin{itemize}[leftmargin=3mm]
                \item 1. the output is generic or completely irrelevant, e.g., \textit{``I don’t think it is untrue to believe in such an assumption.''}

                \item 3. the output mentions at least one major overlapping concepts to the topic, e.g., \textit{``Arguments for raising minimum wage should be the same as arguments for universal basic income. It will result in higher prices eventually.''}

                \item 5. the output is on topic and has a clear opposing stance, e.g., \textit{``The minimum wage was designed specifically for someone to be able to pay rent, utilities, and food. However, this standard is not met in most states.''}
                \end{itemize}
        
        \end{itemize}} 
         \\
         \hline
    \end{tabular}
    \caption{Evaluation guidelines on the first human study and representative examples on rating scales.}
    \label{tab:human-eval-1}
\end{figure*}

\begin{figure*}[h]
    \centering\fontsize{9}{11}\selectfont
    \begin{tabular}{|p{135mm}|}
         \hline
         This study aims to compare some intervention strategies over the same model. The same 53 entries from study 1 will be evaluated. Additionally, 53 entries from a different domain (opinion) will be presented. For each entry, please read the title and three samples, then determine an overall rank, from best to the worst, e.g., ``\texttt{B > A > C}'' to indicate \texttt{B} is better than \texttt{A}, and \texttt{A} is better than \texttt{C}. We allow ties among samples if such order is difficult to decide, in which case please indicate with a ``\texttt{-}'' symbol, such as ``\texttt{B - A > C}''.
         \\ \quad \\
         
         When evaluating the examples, please consider fluency, coherence, and semantic correctness (e.g., whether the sample contradicts common sense, or misuses subjective/objective for certain predicates). An example is shown below:

        {\fontsize{9}{11}\selectfont
        \begin{itemize}[leftmargin=3mm]
            \item {\bf Entry:} The Dog in My Life (and Maybe a Cat, Too)
            {\fontsize{9}{11}\selectfont
            \begin{itemize}[leftmargin=3mm]
                \item \texttt{A}: \textit{Thank you to John Tierney shedding light on the many health problems endured by adopted cats and dogs. I have my share of those problems and have recently adopted a purebred Himalayan cat. His concave face and abdomen makes breathing difficult, and none of the hairs on his ears are watted back, and resulting in irritation. He also has fur mats like cotton candy and requires constant clipping to keep him from being dirty and smeared. I told that some Himalayan owners consider him a "lion's skin." I am happy to say he is not, though I don't know why. I did not need anesthesia to remove all the fur from his face, tail and paws. Clearly, my adopted cat is healthy and is a lot of fun to be around. I am not afraid of him being around the house alone.}
                
                \item \texttt{B}: \textit{I thank Maureen Dowd for shedding light on the many health problems of Himalayans. I not, however, read the article about those problems when she first adopted a purebred Himalayan cat. Her concave face, which makes breathing difficult, and none of the hairs around her eyes are allowed to stand up, resulting in irritation and scratching. Her fur mats like cotton candy and requires constant clipping (even on the off-duty days). I have been told that some Himalayan owners think that their pet would be a "lion" or a "elephant," but I can't tell her that. The treatment, which I am told, is anesthesia and applies to the fur on her face, tail and paws. Like my first adopted cat, I was told if I didn't to stand in front of the electric chair. Instead, let her meow Iona.}
                
                \item \texttt{C}: \textit{Thank you to John Tierney shedding light on the many health problems facing dogs and cats in America. I can add one of my those problems: I recently adopted a purebred Himalayan cat. His concave face and body makes breathing difficult and the back of the hairs behind his ears are pulling down, pulling and resulting in irritation. He also has fur mats like cotton candy and requires constant clipping, which can be painful and painful. I have told that some Himalayan owners put on a "lion's skin." If the owner decides that this is correct, then he is right, and I should be prepared to anesthesia to remove all the fur from my face, tail and paws. Luckily, my adopted cat is extremely intelligent, which means that he would understand me if I asked him to leave me alone. We live in a very close community. }
            \end{itemize}}
            
            \item \texttt{>>> overall ranking: A > B - C}
            \item Reasons: \texttt{A} is generally grammatical and coherent. \texttt{C} has some grammatical errors such as \textit{``I can add one of my those problems''} and repetitions such as \textit{``pulling down, pulling and\ldots''}, \textit{``which can be painful and painful''}, and some semantic problems such as \textit{``remove all the fur from my face, tail and paws''}.  \texttt{B} has grammatical errors such as \textit{``I not, however, read\ldots''}, \textit{``if I didn’t to stand\ldots''}, and some semantic problems, such as \textit{``Her fur mats like cotton candy and requires constant clipping''} and \textit{``stand in front of the electric chair''}.

            (Annotators do not need to provide reasons for the rank.)
            
        \end{itemize}} 
         \\
         \hline
    \end{tabular}
    \caption{Evaluation guidelines on the second human study and annotation examples.}
    \label{tab:human-eval-2}
\end{figure*}

\section{Sample Output}
\label{sec:sample}
From Figure~\ref{tab:arggen-sample-1} to Figure~\ref{tab:news-sample-2} we show more sample outputs, together with their human-written references.

\begin{figure*}[h]
	\fontsize{10}{11}\selectfont
    \centering
	\begin{tabular}{|p{8mm}p{140mm}|}
    \hline
	\multicolumn{2}{|l|}{\textbf{Prompt}: CMV : The EU need its own independent armed forces } \\
	\rowcolor{lightgray!30}
	\multicolumn{2}{|l|}{\textsc{Human}:} \\
     & I wonder what you think the primary benefit of a military under the authority of EU institutions would be. It strikes me that the EU is already able to deploy its armed forces without UN Security Council authorisation - - the EU member states, or any subset of EU member states, can decide to do so. Do you think the primary advantage of a EU military is that military operations conducted under the authority of EU institutions would have a level of legitimacy that neither individual national states ( or coalitions of such states ) or the UN security council could provide? If so, can you say more about why you think that? \\
	\rowcolor{lightgray!30}
	\multicolumn{2}{|l|}{\textsc{KPSeq2seq}:} \\
	& Well, if the EU is armed forces it has the authority. If the EU is armed forces it has the authority. Even the "authority" does not give them that advantage. If it does they are just not armed. \\
	\rowcolor{lightgray!30}
	\multicolumn{2}{|l|}{\textsc{PAIR}$_\text{light}$:} \\
	 & To me, the primary benefit of having a military is the authority of EU institutions. By having the EU armed forces, you have UN Security Council authorisation from the EU member states ( or any subset of EU member states, depending on who you ask ). I think the primary advantage of having a EU military is a level of legitimacy to have your own, your military operations conducted under the authority of EU institutions, and a level of legitimacy that it simply would not exist if they are conducted under the authority of individual national states or coalitions such as most states in the UN security council. 2 \\
	\rowcolor{lightgray!30}
	\multicolumn{2}{|l|}{\textsc{PAIR}$_\text{full}$ w/o refine:} \\
	 &  I don't see the primary benefit of a military under the authority of EU institutions. Personally, I don't think the EU is much different from the UN armed forces... UN Security Council authorisation applies to the EU member states, but any subset of EU member states that is not fully armed cannot. Also, the primary advantage of a EU military is that military operations conducted under the authority of EU institutions still have a level of legitimacy which allows individual national states, or coalitions of such states, to the UN security council, so it's less likely to fall under the less - delegated. So, I would argue that it \\
	\rowcolor{lightgray!30}
	\multicolumn{2}{|l|}{\textsc{PAIR}$_\text{full}$:} \\
	& I don't see the primary benefit of a military under the authority of EU institutions. Instead, I think the EU should set up it's own armed forces with UN Security Council authorisation, to the EU member states, or any subset of EU member states that wishes not to participate. Ultimately, the primary advantage of a EU military is that military operations conducted under the authority of EU institutions already have a level of legitimacy comparable to individual national states, or coalitions of such states, to the UN security council. I think there's likely to be less the other way around but likely to be more the other. )
 \\
	\hline
	\end{tabular}
	\caption{
	Sample outputs on argument generation.
  }\label{tab:arggen-sample-1}
\end{figure*}

\begin{figure*}[t]
	\fontsize{10}{11}\selectfont
    \centering
	\begin{tabular}{|p{8mm}p{140mm}|}
    \hline
	\multicolumn{2}{|l|}{\textbf{Prompt}: CMV : We shouldn't raise the minimum wage to \$ 15 an hour or something similar. } \\
	\rowcolor{lightgray!30}
	\multicolumn{2}{|l|}{\textsc{Human}:} \\
     & People need to live. They need to be able to afford basic things like food, housing, education, transportation and clothes. Currently, that isn't possible with most minimum wages in the US. It's disgraceful that businesses are allowed to let workers have one or more full time jobs without paying them enough to live a very basic life. Increasing the wages will decrease the profits of the company. But who does that hurt? In most cases it's the really wealthy owners, the CEO's and others in the top 1\%. Is that really a bad thing? I'm not saying that they shouldn't be allowed to make lot of money, but do you really believe that they should be able to make that money at the cost of the people they employ? If the businesses aren't able to survive while paying living wages to their workers, maybe they shoud give the place to someone that can? \\
	\rowcolor{lightgray!30}
	\multicolumn{2}{|l|}{\textsc{KPSeq2seq}:} \\
	& So why are there cases where wages will skyrocket, and not people who just do basic things. Why not? People should be able to have their wages grow and if you say that that money is the best then yes. But if that is not what you're arguing then it doesn't say much. \\
	\rowcolor{lightgray!30}
	\multicolumn{2}{|l|}{\textsc{PAIR}$_\text{light}$:} \\
	 & People want to earn as much as possible. They want to be able to afford basic things like food, housing, education, transportation, clothes, and so on. If minimum wages go up they will have to go up. Increasing the wages will decrease the profits of the company. In most cases these tend to be wealthy owners who are often the CEO's of the top 1\% of the US. This is a bad thing. Because if you make lot of money, you make that money at the cost of the people around you. So the businesses need to be willing to pay living wages to their workers and give the place to it's workers. \\
	\rowcolor{lightgray!30}
	\multicolumn{2}{|l|}{\textsc{PAIR}$_\text{full}$ w/o refine:} \\
	 &  People should be given as little as it takes to afford basic things like food, housing, education, transportation, clothes, etc. However, when you think minimum wages should never be set. It's disgraceful for businesses to discriminate against their workers and work in more full time jobs, where someone has to pay for their basic life. Increasing the wages will decrease the profits for the company where it hurts most. And in most cases it's the extremely wealthy owners or the CEO, not necessarily about the top 1\%. But this is a bad thing for everyone. You have to be willing to set aside somewhere between make lot's and make that money, but the cost is not all the \\
	\rowcolor{lightgray!30}
	\multicolumn{2}{|l|}{\textsc{PAIR}$_\text{full}$:} \\
	& People can work as much as it takes to afford basic things like food, housing, education, transportation, clothes, etc. So, that is what minimum wages should be based on. It is disgraceful for businesses to discriminate against their workers because to get more full time jobs is not for them to live for their basic life. Increasing the wages will decrease the profits of the company they pay them for. In most cases, they do have wealthy owners - the CEO may be already in the top 1\%. And that is a bad thing. It is much harder to find something to live for. Now they make lot more to make that money, but the cost is higher for the people )
 \\
	\hline
	\end{tabular}
	\caption{
	Sample outputs on argument generation.
  }\label{tab:arggen-sample-2}
\end{figure*}

\begin{figure*}[t]
	\fontsize{10}{11}\selectfont
    \centering
	\begin{tabular}{|p{8mm}p{140mm}|}
    \hline
	\multicolumn{2}{|l|}{\textbf{Prompt}: Putting Profit Above Health } \\
	\rowcolor{lightgray!30}
	\multicolumn{2}{|l|}{\textsc{Human}:} \\
     & Re ``Chicken With Arsenic? Is That O.K.?'' (Eating Well, April 5): Chicken with arsenic is most certainly not O.K. Most chickens raised for meat in the United States continue to be fed this poison, making most chicken meat at risk for contamination. It says a lot about the filthy, overcrowded conditions in which most factory-farmed chickens are confined that the addition of antibiotics or antiparasitic compounds like arsenic can result in such profitable increases in growth. By selectively breeding chickens for healthier immune systems instead of just for plumper meat, the industry can move away from risky crutches like arsenic. Especially now, with the threat of avian influenza, the industry can no longer be allowed to let its bottom line trump human or animal health. \\
	\rowcolor{lightgray!30}
	\multicolumn{2}{|l|}{\textsc{KPSeq2seq}:} \\
	& Re ``The Real Face of Chicken'' (editorial, Feb. 16): A major health hazard of factory-farmed chickens raised for meat comes from contamination of meat and poultry with arsenic, pesticides and hormones. Arsenic is made by breeding chickens in the presence of antiparasitic compounds. As an alternative, people with higher blood levels may have healthier immune systems. People with high cholesterol may be able to eat more plumper meat. People with high blood pressure may be able to eat more of their own blood. The only question in this whole controversy is whether the animal farming industry values animal health over human health. For the vast majority of chickens raised for meat, the health risks are less than if they had been raised in natural settings instead of in the filthy, overcrowded conditions of factory farms. Even with the most sophisticated new technology, the industry continues to use antibiotics and heavy metal, making the supply of chicken meat much more vulnerable to contamination, including by avian influenza. It also puts people on risky crutches. \\
	\rowcolor{lightgray!30}
	\multicolumn{2}{|l|}{\textsc{PAIR}$_\text{light}$:} \\
	 & Re ``Chicken With Arsenic? It Defends itself,'' by Michael O'Hanlon (Editorial Observer, April 2): Chicken with arsenic and cancer-causing, too, poses threat to humans by being raised for meat, making one of the world's largest source of chicken meat contamination. The F.D.A.'s recent report on meat-producing farms shows that at least half of the cases in the United States -- the filthy, overcrowded conditions of factory-farmed chickens -- have been related to diseases that range to the use of antibiotics and antiparasitic compounds like arsenic, which causes cancer in the body. Yet, in the name of cutting costs and breeding chickens with healthier immune systems and plumper meat, the industry has put the healthiest members of its ranks on, risky crutches, on the line -- and, in this case, in the case of China, exposing them to arsenic. This is a practice that is a source of profit for the industry and, such as avian influenza, the industry's bottom line. \\
	\rowcolor{lightgray!30}
	\multicolumn{2}{|l|}{\textsc{PAIR}$_\text{full}$ w/o refine:} \\
	 & Re ``Chicken Arsenic? Not Really'' (editorial, March 13): Your editorial raises several important Chicken Against arsenic issues. First, the fact that millions of chickens raised for meat are slaughtered each year means that there are millions of pieces of chicken meat that are without contamination. Second, the effects of the filthy, overcrowded conditions on factory and factory-farmed chickens mean that the natural addition of antibiotics and antiparasitic compounds to arsenic is unnecessary. And third, in addition to raising and breeding chickens with healthier immune systems and producing healthier and plumper meat, the industry also uses the equally risky crutches of arsenic. When it comes to the spread of avian influenza, the industry's bottom line should be a bottom-line decision, not an ethical health one.  \\
	\rowcolor{lightgray!30}
	\multicolumn{2}{|l|}{\textsc{PAIR}$_\text{full}$:} \\
	& Re ``Chicken Arsenic? Get Out!'' (editorial, April 21): Your editorial is right. Chicken has arsenic in its blood, and given that millions of chickens raised for meat are slaughtered every year, that is a large portion of all of chicken meat's sources of contamination. Second, another consequence of the filthy, overcrowded conditions of factory and factory-farmed chickens is that they are given antibiotics and antiparasitic compounds, arsenic and antibiotics. And third, by trying to save money and breeding chickens with healthier immune systems in the name of plumper meat, the industry is surviving on the risky crutches of arsenic. When it comes to the fear of avian influenza, the industry's bottom line should be bottom-line profits, not public health and safety. Chicago, April 23, 2005 \\
	\hline
	\end{tabular}
	\caption{
	Sample outputs on opinion generation.
  }\label{tab:opinion-sample-1}
\end{figure*}

\begin{figure*}[t]
	\fontsize{10}{11}\selectfont
    \centering
	\begin{tabular}{|p{8mm}p{140mm}|}
    \hline
	\multicolumn{2}{|l|}{\textbf{Prompt}: The World's Best Job } \\
	\rowcolor{lightgray!30}
	\multicolumn{2}{|l|}{\textsc{Human}:} \\
     & I read with interest your article about travel writers. While I'm sure that those travel writers who research guides to Goa and Romania have a rough time, the article really doesn't paint a complete picture of all travel writers. It's true that most of us don't make a fortune on our books and that research trips can be hectic, but some of us clearly do have the ideal job or hobby. I am an author of ``Eating \& Drinking in Paris'' and ``Eating \& Drinking in Italy,'' among other travel guides. Can you imagine a better job than researching restaurants, bars and cafes in France and Italy? \\
	\rowcolor{lightgray!30}
	\multicolumn{2}{|l|}{\textsc{KPSeq2seq}:} \\
	& Regarding the article about travel writers [``Drinking, Eating and Pining'' in April 15]: I've found that my hobby is researching restaurants and cafes. It is the ideal job for a traveler. I have been doing research trips to Goa, Greece, Romania, Italy, South America and Europe, among other travel guides. That's why I think all travel writers should get back on their feet. Being a part owner, traveling writer, my partner and I have traveled to India and have taken our sons and daughters on trips. I can't imagine a better job for me. If I have a rough time with time management, just write and I'll get it. My job does require overtime, but my life is not hectic enough. And the best part is that I can share my stories with all those travel writers who write about their trips. Travel writers should ask about research guides and why they write about it. \\
	\rowcolor{lightgray!30}
	\multicolumn{2}{|l|}{\textsc{PAIR}$_\text{light}$:} \\
	 & I would like to thank Kerry Brown who reminded us what a wonderful part of the world it is for travel writers and the part of the world that requires so much of us. While she is probably one of those travel writers who "research and I," research guides in Goa, Romania or anywhere else she chooses, the travel writer has a rough time finding a home for all travel writers no matter where we are or how to do it. And for the most part, when we are not at home we do our research trips in a very short period of time, which is hectic and mind-nanding, but it is still the ideal job and hobby for us. Ms. Brown's article was timely, because one of the offices of the Food and Drug Administration, which is mentioned in the article as a source, wrote about a guide, ``Eating and Drinking in Italy, along with the work of other travel guides, on how to eat well and drink there.'' I cannot imagine a better job than researching restaurants and cafes in Italy, as it would be to do so every day if I traveled to places as far as Italy and as far as parts of the world. \\
	\rowcolor{lightgray!30}
	\multicolumn{2}{|l|}{\textsc{PAIR}$_\text{full}$ w/o refine:} \\
	 &   It was good to see about travel writers. I am a little sorry those travel writers and research guides from Goa and Romania had a rough time there, but there is a whole industry, which is all travel writers. They work very hard, very well, they can travel around the world (and do research trips that are hectic), and they are certainly more than the ideal job and hobby. I was very happy to find ``Eating and Drinking in Italy,'' by Pia Portora Drinking in Italy, and other travel guides. Can you imagine a better job than researching restaurants, stores and cafes in Paris and Italy?.  \\
	\rowcolor{lightgray!30}
	\multicolumn{2}{|l|}{\textsc{PAIR}$_\text{full}$:} \\
	& I was happy to see that travel writers are getting better. Many of those travel writers and research guides in Goa and Romania had a rough time there, but the vast majority of them, as is all travel writers. They all work hard and do a lot of research trips all over the world. They research trips, are hectic, and they are much more than the ideal job or hobby. I was particularly struck by the ``Eating and Drinking in Italy,'' by Maliafra Drinking in Italy, and other travel guides. Can you imagine a better job than researching restaurants, hotels and cafes in places like Italy? I can. \\
	\hline
	\end{tabular}
	\caption{
	Sample outputs on opinion generation.
  }\label{tab:opinion-sample-2}
\end{figure*}

\begin{figure*}[t]
	\fontsize{10}{11}\selectfont
    \centering
	\begin{tabular}{|p{8mm}p{140mm}|}
    \hline
	\multicolumn{2}{|l|}{\textbf{Prompt}:  McGarry Bowen Added To Kraft Agency Roster} \\
	\rowcolor{lightgray!30}
	\multicolumn{2}{|l|}{\textsc{Human}:} \\
     & Kraft Foods is expanding its agency roster by naming McGarry Bowen in New York to create campaigns for salad dressings, mayonnaise and barbecue sauces bearing the Kraft brand name. Spending was estimated at \$30 million to \$40 million. The hiring of McGarry Bowen is part of a trend on Madison Avenue, as big marketers like Kraft shift creative assignments to smaller, independent agencies from larger agencies owned by holding companies. The Chicago office of JWT, part of the WPP Group, had created ads for Kraft salad dressings, while the Chicago office of Foote Cone \& Belding, part of the Interpublic Group of Companies, had created ads for the other products. Both JWT and Foote Cone remain on the Kraft agency roster. \\
	\rowcolor{lightgray!30}
	\multicolumn{2}{|l|}{\textsc{KPSeq2seq}:} \\
	& Re ``Kraft to hire an ad agency'' (front page, Jan. 12): Not only has the Chicago office of JWT, which is part of Kraft Foods, made an important contribution to the campaign to take the campaign off the ground, but it is also part of an industry trend that has seen major companies like JWT shift creative assignments to independent agencies. Both JWT and Kraft remain on the Kraft agency roster. The Chicago office has not changed its agency roster, so it is not part of the process. In fact, in the last month, we have created ads for potato chips, salad dressings and barbecue sauces that have never before been used by anyone bearing the Kraft brand name in an advertising campaign. It is also one of the last major, publicly traded advertising campaigns to be financed by one of the nation's big marketers. Instead of naming McGarry Bowen to represent us, McGarry Bowen is now being hired by one of our larger agencies, Foote Cone of New York. McGarry Bowen is one of the world's largest marketers of Kaffeeklatsch, Belding and other soft-drink brands and the maker of Kraft salad dressings and mayonnaise. \\
	\rowcolor{lightgray!30}
	\multicolumn{2}{|l|}{\textsc{Ours}$_\text{order}$: + refine} \\
	 & Kraft Foods in Cincinnati is expanding its agency roster by naming McGarry \& Bowen in Chicago to create ads for its salad dressings, mayonnaise and even barbecue sauces bearing the Kraft brand name. Spending was estimated at more than \$200 million. McGarry \& Bowen's hiring is part of a trend by big marketers like Kraft to shift creative assignments from independent agencies to larger agencies. The Chicago office of JWT, part of the WPP Group, created ads for Kraft's ``soup'' dressings and mayonnaise as well as those for other products; the Chicago office of Foote Cone \& Belding, part of the Interpublic Group of Companies, created ads for "soup" dressings, for which Kraft is a producer of record, as well as for General Mills. Both JWT, part of the Omnicom Group, and Foote Cone still remain on the Kraft agency roster. The decision comes after Foote Cone, part of Foote Cone \& Be, agreed to become a consultant to McGarry \& Bowen on the account. Both McGarry \& Bowen, part of the Omnicom Group, and J. Walter Thompson Worldwide in New York, and WPP Worldwide Advertising of Kansas City, Mo., continue to create ads for Kraft. \\
	\rowcolor{lightgray!30}
	\multicolumn{2}{|l|}{\textsc{PAIR}$_\text{full}$ w/o refine:} \\
	 & Kraft Foods expanded its agency roster by naming McGarry Bowen in New York to create campaigns for salad dressings, mayonnaise and barbecue sauces bearing the Kraft brand name. Spending was estimated at \$20 million. The decision to open an agency with McGarry Bowen is part of a trend in which large clients like big marketers like Kraft shift creative assignments from large, independent agencies to larger agencies in the same market. The Chicago office of JWT, part of the WPP Group, had created ads for Kraft salad dressings, and the Chicago office of Foote Cone \& Belding, part of the Interpublic Group of Companies, had created ads for its ice cream. Both JWT and Foote Cone remain on the Kraft agency roster. \\
	\rowcolor{lightgray!30}
	\multicolumn{2}{|l|}{\textsc{PAIR}$_\text{full}$:} \\
	& Kraft Foods expanded its agency roster by naming McGarry Bowen in New York to create campaigns for salad dressings, mayonnaise and barbecue sauces bearing the Kraft brand name. Spending was estimated at \$10 million. The decision to expand its relationship with McGarry Bowen is part of a trend in which big marketers like big marketers like Kraft shift creative assignments from small, independent agencies to larger agencies at the same time. The Chicago office of JWT, part of the WPP Group, had created ads for Kraft salad dressings, while the Chicago office of Foote Cone \& Belding, part of the Interpublic Group of Companies, had created ads for ice cream. Both JWT and Foote Cone remain on the Kraft agency roster.\\
	\hline
	\end{tabular}
	\caption{
	Sample outputs on news generation.
  }\label{tab:news-sample-1}
\end{figure*}

\begin{figure*}[t]
	\fontsize{10}{11}\selectfont
    \centering
	\begin{tabular}{|p{8mm}p{140mm}|}
    \hline
	\multicolumn{2}{|l|}{\textbf{Prompt}:  Two Held in Man's Death} \\
	\rowcolor{lightgray!30}
	\multicolumn{2}{|l|}{\textsc{Human}:} \\
     & A 17-year-old girl and a 16-year-old boy were arraigned as adults in State Superior Court here yesterday in the stabbing death of a 34-year-old New Britain man attacked outside his home earlier this week. The teenagers, Elsie Figueroa, 17, and Luis Gabriel Rivera, 16, did not enter pleas to the charges of murder and conspiracy to commit murder, and each was being held in \$1 million bail. In addition, a 13-year-old boy was charged as a juvenile in the killing and was scheduled to appear in juvenile court on Monday. His name was withheld because of his age. On Wednesday night, the New Britain man, Jaroslaw Sawicki, was found bleeding from stab wounds to the chest near the home he shared with his parents, the authorities said. \\
	\rowcolor{lightgray!30}
	\multicolumn{2}{|l|}{\textsc{KPSeq2seq}:} \\
	& I am a 16-year-old boy, whose brother was murdered at the age of 13, a victim of Elsie Figueroa's violence in 1987, and who will be charged as a juvenile. I was one of the people at the scene when Luis Gabriel Rivera stabbed his first wife, Jaroslaw Sawicki, to death at the age of 14, and I was the only witness to a 34-year-old New Britain man's stabbing death, with all of the help of a 13-year-old boy, who was bleeding from stab wounds. The teenagers charged are all in custody, and they have a good chance of not getting to see the New Britain man again. The young people charged, who have no previous criminal record, did not commit murder, and they should not be held accountable in juvenile court. Instead, they should be allowed to enter pleas of guilty, and they should be given the benefit of the doubt. \\
	\rowcolor{lightgray!30}
	\multicolumn{2}{|l|}{\textsc{PAIR}$_\text{light}$:} \\
	 & The parents of a 16-year-old boy charged with manslaughter in the stabbing death of a 34-year-old New Britain man were arrested. The teenagers were arraigned in Superior Court here on Thursday charges that they and two of the parents, Elsie Figueroa, and her son, identified as Luis Gabriel Rivera, were arraigned and forced to enter pleas to two counts of murder in the first degree, one conspiracy to commit murder and weapons possession. Mr. Castillo's father, Antonio Castillo Jr., who is also the father of a 13-year-old boy, was not released here as he was arraigned and charged as a juvenile in juvenile court. The indictments, which were reported Friday in The News of Westport, were made by the Suffolk County prosecutor's office. The authorities say they believe the New Britain man, Jaroslaw Sawicki, was stabbed six times in the head. The police said the cause of death of Mr. Castillo, who was bleeding from stab wounds, has not been determined but that surgically, Sawicki was dead by the time. Mr. Castillo's lawyer, Robert R. Gross, said, "We don't know anything that indicates that he was ever stabbed" and denied that his client was stabbed because of his age. "Our client is continuing to fight," Mr. Gross said. "We are going to continue to try to get to the bottom of this of the whole thing." \\
	\rowcolor{lightgray!30}
	\multicolumn{2}{|l|}{\textsc{PAIR}$_\text{full}$ w/o refine:} \\
	 & The mother of a suspect here and a 16-year-old boy were arraigned yesterday on charges that they took part in the fatal stabbing death of a 34-year-old New Britain man on Nov. 3, officials said. The teenagers, Elsie Figueroa, 17, and Luis Gabriel Rivera, were not required to enter pleas; the two were charged with conspiracy to commit murder, a felony. Another suspect in the case, Jose Rodriguez, 17, a 13-year-old boy, charged as a juvenile, is to be tried as an adult in juvenile court on Nov. 15. All three are free on bond. The police said that the New Britain man, Jaroslaw Sawicki, was found bleeding from stab wounds to the neck and head. He was pronounced dead on Nov. 3 at \\
	\rowcolor{lightgray!30}
	\multicolumn{2}{|l|}{\textsc{PAIR}$_\text{full}$:} \\
	& A pair of teenagers, one and a 16-year-old boy, were arraigned yesterday on charges that they took part in the stabbing death of a 34-year-old New Britain man on Dec. 18, officials said. The teenagers, Elsie Figueroa, 17, and Luis Gabriel Rivera, 16, declined to enter pleas for themselves and were charged with conspiracy to commit murder, a felony. Also yesterday in the case, the police, said, a 13-year-old boy was charged as a juvenile, and will likely be tried as an adult in court. Ms. Figueroa was released on bond. The police said that the New Britain man, Jaroslaw Sawicki, was found bleeding from stab wounds in the basement of his home and was pronounced dead at St. Joseph. \\
	\hline
	\end{tabular}
	\caption{
	Sample outputs on news generation.
  }\label{tab:news-sample-2}
\end{figure*}

\end{document}